\documentclass[11pt,a4paper]{article}
\usepackage[hyperref]{eacl2021}
\usepackage{times}
\usepackage{latexsym}
\pdfoutput=1

\usepackage{microtype}
\usepackage{amsmath}
\usepackage{multirow}
\usepackage{graphicx}
\usepackage{makecell}
\usepackage{stfloats}

\newcommand\rurl[1]{%
	\href{https://#1}{\nolinkurl{#1}}%
}

\aclfinalcopy 
\def\aclpaperid{1043} 

\title{Entity-level Factual Consistency of Abstractive Text Summarization}

\author{Feng Nan$^1$ \quad Ramesh Nallapati$^1$ \quad Zhiguo Wang$^1$ \quad Cicero Nogueira dos Santos$^1$ \\ \quad \textbf{Henghui Zhu}$^1$ \quad \textbf{Dejiao Zhang}$^1$ \quad \textbf{Kathleen McKeown}$^{1,2}$ \quad \textbf{Bing Xiang}$^1$ \\ Amazon Web  Services$^1$, Columbia University$^2$\\
		{ \{nanfen, rnallapa, zhiguow, cicnog, henghui, dejiaoz, mckeownk, bxiang\}@amazon.com}
}

\date{}

\begin{document}
\maketitle
\begin{abstract}
A key challenge for abstractive summarization is ensuring factual consistency of the generated summary with respect to the original document.
For example, state-of-the-art models trained on existing datasets exhibit entity hallucination, generating names of entities that are not present in the source document.
We propose a set of new metrics to quantify the entity-level factual consistency of generated summaries and  we show that the entity hallucination problem can be alleviated by simply filtering the training data.
In addition, we propose a summary-worthy entity classification task to the training process as well as a joint entity and summary generation approach, which yield further improvements in entity level metrics.
\end{abstract}

\section{Introduction}
Many recent advances in deep neural networks have led to significant  improvement in the quality of abstractive summarization \cite{radford2019language, gehrmann-etal-2019-generating, lewis2019bart}.
Despite this progress,
there are still many limitations facing neural text summarization \cite{kryscinski-etal-2019-neural},
the most serious of which is their tendency to generate summaries that are not factually consistent with the input document; 
a factually consistent summary only contains statements that can be derived from the source document. Recent studies show that about 30\% of the summaries generated by neural network sequence-to-sequence models suffer from fact fabrication \cite{AAAI1816121}. Unfortunately, the widely used ROUGE score is inadequate to quantify factual consistency \cite{kryscinski-etal-2019-neural}.  

Factual inconsistency can occur at either the entity or the relation level.
At the entity level, a model generated summary may contain named-entities that never appeared in the source document. We call this the entity \emph{hallucination} problem. For example, consider the following model generated summary:
\begin{quotation}
\emph{People in Italy and the Netherlands are more likely to consume fewer cups of coffee than those in the \underline{UK}, a study suggests.}
\end{quotation}
``UK'' never appeared in the input source document (taken from the test set of the XSUM dataset \cite{narayan-etal-2018-dont}). In fact, the source document mentioned a study involving people in Italy and Netherlands; ``UK'' was a result of model hallucination. 
Another type of inconsistency occurs when the entities indeed exist in the source document but the relations between them are not in the source document. This type of inconsistency is much harder to identify.
Open Information Extraction (OpenIE) and dependency parsing tools have been used \cite{AAAI1816121} to identify the underlying relations in a summary, but
are not yet accurate enough for practical use.
Ultimately, these researchers
relied on manually classifying generated summaries into \emph{faithful}, \emph{fake}, or  \emph{unclear}. 

In this paper, we propose a set of simple metrics to quantify  factual consistency at the entity-level.
We analyze the factual quality of summaries produced by the state-of-the-art BART model \cite{lewis2019bart} on three news datasets. We then propose several techniques including data filtering, multi-task learning and joint sequence generation to improve performance on these metrics. We leave the relation level consistency to future work.

\section{Related work}
Large transformer-based neural architectures combined with pre-training have set new records across many natural language processing tasks \cite{NIPS2017_7181, devlin-etal-2019-bert, radford2019language}. In particular, the BART model \cite{lewis2019bart} has shown superior performance in many text generation tasks including abstractive summarization. In contrast to encoder-only pre-training such as in BERT \cite{devlin-etal-2019-bert} or decoder-only pre-training such as in GPT-2 \cite{radford2019language}, 
BART is an encoder-decoder transformer-based neural translation model jointly pre-trained to reconstruct corrupted input sequences of text. 

Several authors have pointed out the problem of factual inconsistency in abstractive summarization models \cite{kryscinski-etal-2019-neural, kryciski2019evaluating, AAAI1816121, welleck-etal-2019-dialogue}. 
The authors in \cite{kryciski2019evaluating} proposed to train a neural network model to classify if a summary is factually consistent with a given source document, similar to a natural language inference task. In the dialogue generation setting, authors in \cite{DBLP:journals/corr/abs-1911-03860} proposed using unlikelihood to surpress logically inconsistent responses. Our work is complementary to such existing approaches as we focus on simple entity-level metrics to quantify and improve factual consistency. Our goal of improving entity-level metrics of summaries is also related to controllable abstractive summarization \cite{fan-etal-2018-controllable}, where a list of named-entities that a user wants to see in the summary can be passed as input to influence the generated summary. In contrast, our goal is to \emph{predict} which entities are summary-worthy while generating the summary that contains them. In this view we are trying to solve a more challenging problem.

\section{Entity-level factual consistency metrics} \label{sec:entity}
We propose three new metrics that rely on off-the-shelf tools to perform Named-Entity Recognition (NER). \footnote{We use Spacy \cite{spacy2}.}
We use $\mathcal{N}(t)$ and $\mathcal{N}(h)$ to denote the number of named-entities in the target (gold summary) and hypothesis (generated summary), respectively. We use $\mathcal{N}(h\cap s)$ to denote the number of entities found in the generated summary that can find a match in the source document. If a named-entity in the summary consists of multiple words, we consider it a match as long as any n-gram of the named-entity can be found in the source document. This is meant to capture the situation where the named-entity can be shortened; for example, ``Obama '' is a match for ``Barack Obama'' and ``Harvard'' is a match for ``Harvard University''. When the match is at the unigram level, we make sure that it is not a stop word such as ``the''. We also make the match case-insensitive to accommodate casing variances. 

\paragraph{Precision-source:}
We propose precision-source ($\mathbf{prec}_s$) to quantify the degree of hallucination with respect to the source: $\mathbf{prec}_s = \mathcal{N}(h\cap s) / \mathcal{N}(h).$
It is simply the percentage of named-entities in the summary that can be found in the source. Low $\mathbf{prec}_s$ means hallucination is severe.

We first evaluate the $\mathbf{prec}_s$ score on the ground truth summaries of the 3 datasets: Newsroom \cite{newsroom}, CNN/DailyMail \cite{nallapati-etal-2016-abstractive} and XSUM \cite{narayan-etal-2018-dont}. 
\begin{table}[h]
	\centering
	\resizebox{\columnwidth}{!}{%
	\begin{tabular}{|l|l|l|l|l|l|l|l|l|l|}
		\hline
		\multirow{2}{*}{} & \multicolumn{3}{c|}{Newsroom}                                                     & \multicolumn{3}{c|}{CNNDM}                                                        & \multicolumn{3}{c|}{XSUM}                                                         \\ \hline
		& \multicolumn{1}{c|}{train} & \multicolumn{1}{c|}{val} & \multicolumn{1}{c|}{test} & \multicolumn{1}{c|}{train} & \multicolumn{1}{c|}{val} & \multicolumn{1}{c|}{test} & \multicolumn{1}{c|}{train} & \multicolumn{1}{c|}{val} & \multicolumn{1}{c|}{test} \\ \hline
		avg. $\mathcal{N}(t)$   & 2.08                    & 2.10                  & 2.09                   & 4.36                    & 5.09                  & 4.87                   & 2.08                     & 2.06                  & 2.08                   \\ \hline
		avg. $\mathcal{N}(t \cap s)$ & 1.88                    & 1.90                  & 1.90                   & 4.21                     & 4.92                  & 4.70                   & 1.64                    & 1.64                  & 1.64                   \\ \hline
		$\mathbf{prec}_s$ (\%) & 90.6                    & 90.6                  & 90.5                   & 96.5                    & 96.7                   & 96.6                   & 79.0                     & 79.5                  & 79.3                   \\ \hline
	\end{tabular}
}
	\vspace{-1mm}
	\caption{Average number of named-entities and the $\mathbf{prec}_s$ scores (\%) in the ground truth summary.}
	\label{table:groundTruthSummaryStats}
\end{table}
Table  \ref{table:groundTruthSummaryStats} shows that among the three datasets, the ground truth summaries in XSUM have the lowest $\mathbf{prec}_s$ score.
This is because the ground truth summaries in the XSUM dataset often use  the first sentence of the article as the summary;
the source document is constructed to be the rest of the article and may not repeat the named-entities that appeared in the summary. We hypothesize that the hallucination problem is largely caused by the training data itself. Thus, we propose to perform entity-based data filtering to construct a ``clean'' version of these datasets as described next.
\paragraph{Entity-based data filtering:}
For each dataset, we apply Spacy NER on the gold summary to identify all the named-entities. \footnote{We ignore certain types of entities such as date, time, numerals because they tend to have large variations in representation and are difficult to determine a match in the source document. The appendix contains more details.}
If any of the entities cannot find a match in the source document, we discard the  sentence that contains the entity from the ground truth summary. If the ground truth summary consists of only one sentence and it needs to be discarded, we remove the document-summary pair from the dataset. 
This way, we ensure that our filtered dataset does not contain hallucination of entities ($\mathbf{prec}_s =1$) in the ground truth summary. The dataset size before and after the filtering is shown in Table \ref{table:datasetStats}. About a third of examples are filtered out for XSUM. Again, this is because of the way XSUM dataset is constructed as mentioned in the previous paragraph. As we shall see in Table \ref{table:dataFilterResult}, entity-based data filtering reduces hallucination of the trained model and the effect is especially significant in the XSUM dataset. 
\begin{table*}[h]
	\centering
	\resizebox{\textwidth}{!}{
		\begin{tabular}{|l|r|r|r|r|r|r|r|r|r|}
			\hline
			\multirow{2}{*}{}      & \multicolumn{3}{c|}{Newsroom}                                                     & \multicolumn{3}{c|}{CNNDM}                                                        & \multicolumn{3}{c|}{XSUM}                                                         \\ \cline{2-10} 
			& \multicolumn{1}{c|}{train} & \multicolumn{1}{c|}{val} & \multicolumn{1}{c|}{test} & \multicolumn{1}{c|}{train} & \multicolumn{1}{c|}{val} & \multicolumn{1}{c|}{test} & \multicolumn{1}{c|}{train} & \multicolumn{1}{c|}{val} & \multicolumn{1}{c|}{test} \\ \hline
			\thead{original}               & \thead{922,500 (1.58)}                     & \thead{100,968 (1.60)}                  & \thead{100,933  (1.59)}                    & \thead{287,112 (3.90)}                     & \thead{13,368 (4.13)}                    & \thead{11,490 (3.92)}                     & \thead{203,540 (1.0)}                     & \thead{11,301 (1.0)}                    & \thead{11,299 (1.0)}                     \\ \hline
			\thead{after filtering} & \thead{855,975 (1.62)}                     & \thead{93,678 (1.64)}                    & \thead{93,486 (1.64)}                     & \thead{286,791 (3.77)}                    & \thead{13,350 (3.99)}                    & \thead{11,483 (3.77)}                     & \thead{135,155  (1.0)}                    & \thead{7,639 (1.0)}                     & \thead{7,574 (1.0)}                      \\ \hline
		\end{tabular}
	}
	\vspace{-1mm}
	\caption{Number of examples in three datasets together with the average number of sentences in the ground truth summary (in parentheses) before and after entity-based filtering.}
	\label{table:datasetStats}
\end{table*}

\paragraph{Precision-target and recall-target:} Although the precision-source ($\mathbf{prec}_s$) metric quantifies the degree of entity hallucination with respect to the source document, it does not capture the entity-level accuracy of the generated summary with respect to the ground truth summary. To get a complete picture of the entity-level accuracy of the generated summary, we propose the precision-target ($\mathbf{prec}_t$) score: $\mathbf{prec}_t = \mathcal{N}(h\cap t) / \mathcal{N}(h),$
where $\mathcal{N}(h\cap t)$ is the number of named-entities in the generated summary that can find a match in the ground truth summary; and the recall-target ($\mathbf{recall}_t$) score: $\mathbf{recall}_t = \mathcal{N}(h\cap t) / \mathcal{N}(t),$
where $\mathcal{N}(t)$ is the number of named-entities in the ground truth summary. We compute the F1 score as $F1_t=2 \cdot \mathbf{prec}_t \cdot \mathbf{recall}_t / (\mathbf{prec}_t  + \mathbf{recall}_t)$.
\begin{table*}[ht]
	\centering
	\resizebox{\textwidth}{!}{%
		\begin{tabular}{|l|l|l|l|l|l|l|l|l|l|l|l|l|}
			\hline
			& \thead{training \\ data}   & \thead{Rouge1}  & \thead{Rouge2}  & \thead{RougeL}  & \thead{macro \\ $\mathbf{prec}_s$} &  \thead{micro \\ $\mathbf{prec}_s$} &  \thead{macro \\ $\mathbf{prec}_t$} &  \thead{micro \\ $\mathbf{prec}_t$} &  \thead{macro \\ $\mathbf{recall}_t$} &  \thead{micro \\ $\mathbf{recall}_t$} &  \thead{macro \\ $F1_t$} &  \thead{micro \\ $F1_t$} \\ \hline
			\multirow{4}{*}{Newsroom} & original        & 47.7{\scriptsize $\pm$0.2}    & 35.0{\scriptsize $\pm$0.3} & 44.1{\scriptsize $\pm$0.2} & 97.2{\scriptsize $\pm$0.1}     & 97.0{\scriptsize $\pm$0.1}      & 65.4{\scriptsize $\pm$0.3}     & 62.9{\scriptsize $\pm$0.4}      & 70.8{\scriptsize $\pm$0.3}        & 68.5 {\scriptsize $\pm$0.2}        & 68.0{\scriptsize $\pm$0.2}     & 65.6{\scriptsize $\pm$0.3}    \\ 
			& + filtering & 47.7{\scriptsize $\pm$0.1}   &  35.1{\scriptsize $\pm$0.1} &44.1 {\scriptsize $\pm$0.1} & 98.1{\scriptsize $\pm$0.1}     & 98.0{\scriptsize $\pm$0.0}       & 66.5{\scriptsize $\pm$0.1}       & 63.8{\scriptsize $\pm$0.1}      & 70.2 {\scriptsize $\pm$0.2}        & 67.7{\scriptsize $\pm$0.3}        & 68.3{\scriptsize $\pm$0.1}   & 65.7{\scriptsize $\pm$0.1}   \\ 
			& + classification  & 47.7{\scriptsize $\pm$0.2}   &  35.1{\scriptsize $\pm$0.1}  &  44.2{\scriptsize $\pm$0.2}   &   98.1{\scriptsize $\pm$0.1}     & 98.0{\scriptsize $\pm$0.0}       & 67.2{\scriptsize $\pm$0.4}     & 64.2{\scriptsize $\pm$0.4}     & 70.3{\scriptsize $\pm$0.2}       &  67.8{\scriptsize $\pm$0.4}       & 68.7{\scriptsize $\pm$0.3}    & 65.9{\scriptsize $\pm$0.4}    \\ 
			& ~~ JAENS  & 46.6 {\scriptsize $\pm$0.5}   &  34.3{\scriptsize $\pm$0.3}   &  43.2{\scriptsize $\pm$0.3}   &  {\bf 98.3}{\scriptsize $\pm$0.1}      & {\bf 98.3}{\scriptsize $\pm$0.1}   & {\bf 69.5}{\scriptsize $\pm$1.6}    & {\bf 67.3}{\scriptsize $\pm$1.2}     & 68.9{\scriptsize $\pm$1.5}      &   66.8{\scriptsize $\pm$1.6}        & {\bf 69.2}{\scriptsize $\pm$0.1}    & {\bf 67.0}{\scriptsize $\pm$0.2}     \\ \hline
			\multirow{4}{*}{CNNDM}    & original        & 43.7{\scriptsize $\pm$0.1}& {\bf 21.1}{\scriptsize $\pm$0.1} & 40.6{\scriptsize $\pm$0.1} & 99.5{\scriptsize $\pm$0.1}      & 99.4{\scriptsize $\pm$0.1}       & 66.0{\scriptsize $\pm$0.4}      & 66.5{\scriptsize $\pm$0.4}      & 74.7{\scriptsize $\pm$0.7}        & 75.4{\scriptsize $\pm$0.6}        & 70.0{\scriptsize $\pm$0.2}    & 70.7{\scriptsize $\pm$0.3}    \\ 
			& + filtering & 43.4{\scriptsize $\pm$0.2} & 20.8{\scriptsize $\pm$0.1} & 40.3{\scriptsize $\pm$0.2} & {\bf 99.9}{\scriptsize $\pm$0.0}      &{\bf 99.9}{\scriptsize $\pm$0.0}      & 66.2	{\scriptsize $\pm$0.4}     & 66.6{\scriptsize $\pm$0.3}     & 74.1{\scriptsize $\pm$0.6}        & 74.9{\scriptsize $\pm$0.6}        & 69.9{\scriptsize $\pm$0.2}    & 70.5{\scriptsize $\pm$0.2}     \\ 
			& + classification  &43.5{\scriptsize $\pm$0.2} & 20.8{\scriptsize $\pm$0.2} & 40.4{\scriptsize $\pm$0.2} &  {\bf 99.9}{\scriptsize $\pm$0.0}      &  {\bf 99.9}{\scriptsize $\pm$0.0}     &  {\bf 67.0}{\scriptsize $\pm$0.6}      &  {\bf 67.5}{\scriptsize $\pm$0.5}     &  74.7{\scriptsize $\pm$0.2}       &   75.5{\scriptsize $\pm$0.1}       &  70.6{\scriptsize $\pm$0.3}    &   71.3{\scriptsize $\pm$0.3}    \\ 
			& ~~ JAENS  &42.4 {\scriptsize $\pm$0.6} & 20.2{\scriptsize $\pm$0.2}  & 39.5{\scriptsize $\pm$0.5}  &  {\bf 99.9}{\scriptsize $\pm$0.0}      &  {\bf 99.9}{\scriptsize $\pm$0.0}     &  {\bf 67.9}{\scriptsize $\pm$0.7}       &  {\bf 68.4}{\scriptsize $\pm$0.6}      &  {\bf 75.1}{\scriptsize $\pm$0.7}       &   {\bf 76.4}{\scriptsize $\pm$0.7}        &  {\bf 71.3}{\scriptsize $\pm$0.2}    &  {\bf 72.2}{\scriptsize $\pm$0.2}     \\ \hline
			\multirow{4}{*}{XSUM}     & original        &  {\bf 45.6}{\scriptsize $\pm$0.1} &  {\bf 22.5}{\scriptsize $\pm$0.1} &  {\bf 37.2}{\scriptsize $\pm$0.1} & 93.9{\scriptsize $\pm$0.1}      & 93.6{\scriptsize $\pm$0.2}      & 74.1{\scriptsize $\pm$0.2}      & 73.3{\scriptsize $\pm$0.2}       &  80.1{\scriptsize $\pm$0.1}        &   80.3{\scriptsize $\pm$0.3}       & 77.0{\scriptsize $\pm$0.1}    & 76.6{\scriptsize $\pm$0.2}    \\ 
			& + filtering & 45.4{\scriptsize $\pm$0.1} & 22.2{\scriptsize $\pm$0.1} & 36.9{\scriptsize $\pm$0.1} &  98.2{\scriptsize $\pm$0.0}       &  98.2{\scriptsize $\pm$0.1}      & 77.9{\scriptsize $\pm$0.2}      & 77.3{\scriptsize $\pm$0.2}      & 79.4{\scriptsize $\pm$0.2}        & 79.6{\scriptsize $\pm$0.2}        & 78.6{\scriptsize $\pm$0.1}    & 78.4{\scriptsize $\pm$0.2}    \\ 
			& + classification  & 45.3{\scriptsize $\pm$0.1} & 22.1{\scriptsize $\pm$0.0}  & 36.9{\scriptsize $\pm$0.1} &   98.3{\scriptsize $\pm$0.1}     &  98.2{\scriptsize $\pm$0.1}      &  78.6{\scriptsize $\pm$0.3}      &  {\bf 78.0}{\scriptsize $\pm$0.3}      & 79.5{\scriptsize $\pm$0.3}        & 79.8{\scriptsize $\pm$0.4}        &  {\bf 79.1}{\scriptsize $\pm$0.1}     &  {\bf 78.9}{\scriptsize $\pm$0.1}   \\ 
			& ~~ JAENS  &43.4{\scriptsize $\pm$0.7} & 21.0{\scriptsize $\pm$0.3} & 35.5 {\scriptsize $\pm$0.4}&  {\bf 99.0}{\scriptsize $\pm$0.1}     &  {\bf 99.0}{\scriptsize $\pm$0.1}    &  77.6{\scriptsize $\pm$0.9}      &   77.1{\scriptsize $\pm$0.6}     &  79.5{\scriptsize $\pm$0.6}       &   80.0{\scriptsize $\pm$0.5}       &  78.5{\scriptsize $\pm$0.2}   &  78.5{\scriptsize $\pm$0.1}    \\ \hline
		\end{tabular}%
	}
	\vspace{-1mm}
	\caption{Comparison of models trained using original data, with entity-based data filtering, with an additional classification task and with JAENS. Scores are all in percentages, averaged over 5 runs and shown with standard deviations. We bold the numbers that are significantly better in the sense that the means are separated by at least the standard deviations. We report both the micro and macro averages of our proposed entity-level scores. In all datasets, data filtering leads to higher $\mathbf{prec}_s$ scores, indicating that entity hallucination can be alleviated by this simple technique. In addition, data filtering generally improves other entity level metrics: $\mathbf{prec}_t$, $\mathbf{recall}_t$ and $F1_t$. Adding the classification task (multi-task) or JAENS to data filtering further improves the performance on $\mathbf{prec}_t$ and $\mathbf{recall}_t$  and therefore the overall entity-level $F1_t$.}
	\label{table:dataFilterResult}
		\vspace{-1mm}
\end{table*}

\section{Multi-task learning:}
In addition to entity-based data filtering, we also explore another method to further improve the summarization quality. In particular, we incorporate an additional task of classifying
summary-worthy named-entities in the source document.
A summary-worthy named-entity in the source document is one that appears in the ground truth summary and thus, is a salient entity,  worthy of inclusion in the generated summary. Intuitively, if we can identify these summary-worthy named-entities using the encoder representation, we may potentially increase the  entity-level precision and recall metrics as well as the overall quality of the summary. We achieve this by adding a classification head to the encoder of BART. 
To prepare for the classification label, we first identify the named-entities in the ground truth summary and find the matching tokens in the source document. We then assign the (B)eginning-(I)nside-(O)utside labels to each token of the source document to denote if the token is beginning, inside or outside of a summary-worthy named-entity, respectively. 
During training, we simply add the classification loss for each token at the encoder to the original sequence-to-sequence loss. 

More precisely, let $\{\left(x^i, y^i\right)\}_{i=1}^N$ be a dataset of $N$ examples where $x^i=x^i_1, \dots, x^i_{ts(i)}$ are the tokens of the $i$th source document and $y^i=y^i_1, \dots, y^i_{tt(i)}$ are the tokens of the target (ground truth summary). The standard sequence-to-sequence training minimizes the maximum log likelihood estimation (MLE) loss:
\begin{equation*}
\mathcal{L}^i_{\text{MLE}} (\theta, x^i, y^i) = - \sum_{t=1}^{tt(i)} \log p_{\theta}(y^i_t | x^i, y^i_{<t}).
\end{equation*}
 With summary-worthy entity classification, each example has an additional sequence of BIO labels $z^i=z^i_1, \dots, z^i_{ts(i)}, z^i_t \in \{0,1,2\}$. By adding an additional fully connected layer on top of the BART encoder, we obtain the classification loss 
 \begin{equation*}
 \mathcal{L}^i_{\text{BIO}} (\theta(\text{enc}), x^i, z^i) = - \sum_{t=1}^{ts(i)} \log p_{\theta(\text{enc})} (z^i_t |x^i).
  \end{equation*}
 Finally, we can minimize the joint loss $ \mathcal{L}^i_{\text{Multitask}} = \mathcal{L}^i_{\text{MLE}} + \alpha \mathcal{L}^i_{\text{BIO}},$
  where $\alpha$ is a hyper parameter. We choose $\alpha$ between 0.1 to 0.5 via the validation sets.
  
\section{Joint Entity and Summary Generation:}
We also explore another generative approach to promote entity-level precision and recall metrics. In particular, instead of just generating the summary, we train the BART model to generate the sequence of summary-worthy named-entities, followed by a special token, and then the summary. We call this approach JAENS (Join sAlient ENtity and Summary generation). Similar to the multi-task learning approach discussed earlier, JAENS encourages the model to jointly learn to identify the summary-worthy named-entities while learning to generate summaries. Since the decoder generates the salient named-entities first, the summaries that JAENS generate can further attend to these salient named-entities through decoder self-attention. 

\section{Experiment results}
We use the pre-trained BART-large model in the Fairseq library \cite{ott2019fairseq} to fine-tune on the 3 summarization datasets.\footnote{Our code is available at \rurl{https://github.com/amazon-research/fact-check-summarization}}
The appendix contains additional details of experimental setup.

In Table \ref{table:dataFilterResult}, we
show the effect of the entity-based data filtering. For each dataset, we train two separate models: using the training data before and after entity-based data filtering as shown in Table \ref{table:datasetStats}. We evaluate both models on the ``clean'' test set after entity-based data filtering. We choose this filtered version of the original test set because we only want to measure entity-level consistency against the correct set of entities; using the unfiltered dataset means we could count a hallucinated entity as correct. We observe improvements of $\mathbf{prec}_s$ across all three datasets trained using the filtered subset of data. For example in XSUM, the $\mathbf{prec}_s$ is increased from 93.6\% to 98.2\%, indicating a significant reduction in entity hallucination. In addition, the entity-based data filtering generally improves other entity-level metrics as well. Even with less training data, the entity-based data filtering is able to maintain the ROUGE scores quite well. For XSUM, about 34\% of the training data is filtered out (c.f. Table \ref{table:datasetStats}), which explains the more noticable impact on the ROUGE scores. 
The results in Table \ref{table:dataFilterResult} suggest that entity-level data filtering is a simple yet effective approach to achieve higher entity-level factual consistency as well as general summarization quality. In Table \ref{tab:dataFilterResultQualitative} we provide qualitative examples where the model trained on the original data produces hallucination and the entity-level data filtering removes such hallucination.

Table \ref{table:dataFilterResult} shows that adding the classification task (multi-task) futher increases the $\mathbf{prec}_t$ and $\mathbf{recall}_t$ metric and therefore the overall entity-level $F1_t$ on top of the improvements from data filtering. Similar gains can be observed with JAENS, which out-performs the multi-task approach on CNNDM and Newsroom datasets. The result confirms our intuition that the summaries in JAENS can benefit from attending to the generated salient entities in terms of the entity level metrics. However, the additional complexity during decoding may have hurt the ROUGE scores. 

For the interested readers, we also evaluated the PEGASUS \cite{pmlr-v119-zhang20ae} models for the ROUGE and entity level metrics on these three datasets in the appendix.

\paragraph{Accuracy of entity level metrics:} As our entity level metrics are based on automatic NER tools and heuristics matching rules, errors in both steps can lead to inaccuracy in the metrics. 
By manually checking 10 random ground truth summaries together with the source documents in the validation split of XSUM dataset, we found that all of the named entities are correctly identified by the NER tool and the matchings are correct. Therefore, we believe that even our current NER tool and matching rule already produce high accuracy in practice. 

\begin{table*}[ht]
	\centering
	\resizebox{\textwidth}{!}{%
		\begin{tabular}{p{0.25\linewidth}|p{0.25\linewidth}|p{0.25\linewidth}|p{0.25\linewidth}}
			\thead{Before data filtering} & \thead{After data filtering} &  \thead{With classification} & \thead{Ground truth summary} \\ \hline
			People in Italy and the Netherlands are more likely to consume fewer cups of coffee than those in the \underline{UK}, a study suggests.
			& The desire to drink coffee may be encoded in our DNA, according to scientists.
			& People with a particular gene are more likely to consume fewer cups of coffee, a study has suggested.
			& Researchers have identified a gene that appears to curb coffee consumption. \\ \hline
			A cathedral in \underline{Surrey} is set to be restored after more than £5m was raised to pay for repairs and improvements.
			& A £7m project to save a Grade II-listed cathedral from demolition is set to go ahead. 
			& A cathedral which has been threatened with demolition is set to be saved by a £5m fundraising campaign.
			& A 1960s-built cathedral that was "at serious risk of closure" has raised more than 90\% of its £7m target for urgent repairs and development. \\ \hline
			More than 800,000 chemists in the Indian \underline{capital, Delhi}, have gone on strike in protest against online drug sales.
			& More than 800,000 chemists in India will go on strike on Wednesday to protest against illegal online drug sales.
			& More than 800,000 chemists in India are set to go on strike on Wednesday in a row over the sale of drugs online.
			& At least 800,000 pharmacies in India are on a one-day strike, demanding an end to online drug sales which they say is affecting their business. \\ \hline
			Police officers in \underline{Pembrokeshire} are to be issued with body-worn cameras. & Police officers in Powys are to be issued with body-worn cameras in a bid to improve transparency in the force. 
			& Police officers in Powys are to be issued with body cameras in a bid to improve transparency in the force.
			& A police force has begun the rollout of body cameras for 800 officers and community support officers. \\ \hline
			Wales midfielder \underline{Becky Lawrence} has been speaking to \underline{BBC Sport} about her time as a player-manager with Melbourne City. & It's been a great few weeks for me as a player-manager and now I'm heading home to Wales ahead of the Cyprus Cup. 
			& It's been a very busy few weeks for me as I'm heading home to Wales ahead of the Cyprus Cup.
			& I have certainly had worse 24 hours in my life than winning the Grand Final with Melbourne City and then being named in the Wales squad for the Cyprus Cup.
		\end{tabular}%
	}
	\vspace{1mm}
	\caption{Generated and ground truth summary examples from the test set of XSUM. The first three columns are generated from the model trained without entity-based data filtering, with entity-based data filtering and with the additional classification task, respectively. The right column contains the ground truth summaries. The hallucinated named-entities are underscored. Proposed data filtering overcomes hallucination in these examples.}
	\label{tab:dataFilterResultQualitative}
\end{table*}

\section{Conclusion}
In this paper we study the entity-level factual consistency of the state-of-the-art summarization model. We propose precision-source score $\mathbf{prec}_s$ to quantify the degree of entity hallucination. We also propose additional metrics $\mathbf{prec}_t$ and $\mathbf{recall}_t$ to measure entity level accuracy of the generated summary with respect to the ground truth summary. We found that the ground truth summaries of the XSUM dataset contain a high level of entity hallucination. We propose a simple entity-level data filtering technique to remove such hallucination in the training data. Experiments show that such data filtering leads to significant improvement in $\mathbf{prec}_s$. ($\mathbf{prec}_s$  increases from below 94\% to above 98\% in XSUM for example.) 
We futher proposed a multi-task learning and a joint sequence generation approach to further improve the entity-level metrics.
Overall, combining our proposed approaches significantly reduces entity hallucination and leads to higher entity level metrics with minimal degradation of the ROUGE scores.

\bibliographystyle{acl_natbib}
\bibliography{summarization}

\appendix

\clearpage

\begin{table*}[bp]
	\centering
	\resizebox{\textwidth}{!}{%
		\begin{tabular}{@{}|c|c|c|c|c|c|c|c|c|c|c|c|@{}}
			\hline
			& \thead{Rouge1}  & \thead{Rouge2}  & \thead{RougeL}  & \thead{macro \\ $\mathbf{prec}_s$} &  \thead{micro \\ $\mathbf{prec}_s$} &  \thead{macro \\ $\mathbf{prec}_t$} &  \thead{micro \\ $\mathbf{prec}_t$} &  \thead{macro \\ $\mathbf{recall}_t$} &  \thead{micro \\ $\mathbf{recall}_t$} &  \thead{macro \\ $F1_t$} &  \thead{micro \\ $F1_t$} \\ \hline
			Newsroom & 40.6   & 28.4   & 37.4   & 94.6        & 94.7        & 53.4        & 55.5        & 68.5 & 67.8 & 60.0 & 61.1 \\ \hline
			CNNDM    & 42.5   & 20.7   & 39.6   & 99.1        & 99.0        & 65.9        & 66.7        & 74.7 & 75.7 & 70.0 & 70.9 \\ \hline
			XSUM     & 45.3   & 23.7   & 37.9   & 93.9        & 93.1        & 76.6        & 75.8        & 80.3 & 80.1 & 78.4 & 77.9 \\ \hline
		\end{tabular}%
	}
	\caption{Evaluation of PEGASUS on NER filtered test sets.}
	\label{tab:pegasus}
\end{table*}

\section{Supplementary material for Entity-level Factual Consistency of Abstractive Text Summarization}

\subsection{Details of NER filtering}
We only consider named-entities of the following types: 'PERSON' (People, including fictional.), 'FAC' (Buildings, airports, highways, bridges, etc.), 'GPE' (Countries, cities, states.), 'ORG' (Companies, agencies, institutions, etc.), 'NORP' (Nationalities or religious or political groups.), 'LOC' (Non-GPE locations, mountain ranges, bodies of water.), 'EVENT' (Named hurricanes, battles, wars, sports events, etc.). We ignore other types of entities such as date, time, numerals because they tend to have large variations in representation and are difficult to determine a match in the source document.

\subsection{Details of experimental setup}
We use the pre-trained BART-large model in the Fairseq library \cite{ott2019fairseq} to fine-tune on the 3 summarization datasets. 

In all experiments, we validate the ROUGE scores of the generated summaries on the validation split and early-stop on the epoch with the highest validation score. We use the standard learning rate of 3e-5 for finetuning with linear decay schedule and 500 warmup steps. For Newsroom, we use 4 p3.16xlarge EC2 instances on AWS with a total of 32 Tesla V100 GPUs for finetuning and the effective batch size is 32; for XSUM, we use 1 p3.16xlarge instance with a total of 8 Tesla V100 GPUs and update frequency of 4, giving an effective batch size of 32; for CNNDM, we use 1 p3.16xlarge instance with a total of 8 Tesla V100 GPUs, giving an effective batch size of 8. 

We chose the $\alpha$ parameter for multi-task learning between 0.1 and 0.5 with step of 0.05 based on ROUGE scores on the validation set. We found the best values are 0.3, 0.3 and 0.15 for Newsroom, CNNDM and XSUM, respectively. We observe that the ROUGE and entity level metrics on validation and test sets are very close, with the former slightly higher. 

During decoding, we use beam size of 1 for Newsroom, 4 for CNNDM and 6 for XSUM (to be consistent with the setting in \cite{lewis2019bart}). We did use trigrams blocking in beam search as we did not see much need for this additional step. 

\subsection{Evaluation of PEGASUS \cite{pmlr-v119-zhang20ae}}
In this section we simply evaluate the PEGASUS checkpoints provided by Huggingface \cite{wolf-etal-2020-transformers} on the NER filtered test sets. The checkpoints are downloaded from \url{https://huggingface.co/google/pegasus-newsroom}, \url{https://huggingface.co/google/pegasus-cnn_dailymail} and \url{https://huggingface.co/google/pegasus-xsum}, respectively.
The results are summarized in Table \ref{tab:pegasus}. Note that PEGASUS performances similarly on CNNDM and XSUM but worse on Newsroom compared to BART-large.  

\end{document}